# On Polynomial Sized MDP Succinct Policies

**Paolo Liberatore**                    PAOLO@LIBERATORE.ORG
*Dipartimento di Informatica e Sistemistica*
*Università di Roma "La Sapienza"*
*Via Salaria 113, 00198, Roma, Italy*

## Abstract

Policies of Markov Decision Processes (MDPs) determine the next action to execute from the current state and, possibly, the history (the past states). When the number of states is large, succinct representations are often used to compactly represent both the MDPs and the policies in a reduced amount of space. In this paper, some problems related to the size of succinctly represented policies are analyzed. Namely, it is shown that some MDPs have policies that can only be represented in space super-polynomial in the size of the MDP, unless the polynomial hierarchy collapses. This fact motivates the study of the problem of deciding whether a given MDP has a policy of a given size and reward. Since some algorithms for MDPs work by finding a succinct representation of the value function, the problem of deciding the existence of a succinct representation of a value function of a given size and reward is also considered.

## 1. Introduction

Markov Decision Processes (MDPs) (Bellman, 1957) have been used in AI for planning when the effects of actions are only probabilistically known (Puterman, 1994). The partially observable extension (POMDP) formalizes scenarios in which the observations do not give a complete description of the state.

The best plan in such domains may not be a simple sequence of actions. Indeed, the best action to take may depend on the current state, which is known (partially, for POMDPs) only when the previous actions have been executed. Such conditional plans are named "policies". Finding policies for MDPs is a problem that has been deeply investigated; algorithms have been developed, e.g., value iteration, policy iteration, and methods based on linear programming (Littman, Dean, & Kaebling, 1995). For POMDPs, variants of the value iteration algorithm have been developed (Cassandra, Littman, & Zhang, 1997; Zhang & Zhang, 2001).

Formally, an MDP is composed of a set of states, a set of actions, the specification of the (probabilistic) effects of actions, and a measure of how good a state is considered (reward function). Initially, MDPs were defined in an explicit form: the states are the elements of a given set $\{s_1, \ldots, s_n\}$; what is true or false in each state $s_i$ is not specified. The effects of the actions and the reward function are represented in an explicit form, e.g., if there are $n$ states, a vector of $n$ elements represents the reward function (each element of the vector is the reward of a state.)

While such explicit representation is simple, it is not practical to use in many cases. Indeed, many real-world scenarios can be described by a set of variables (state variables); for example, a set of Boolean variables can specify what is true or false in the current state. An explicit representation of such a domain is always of size exponential in the number of state variables, as it contains an enumeration of all states. This is why succinct representations are used instead: the states are assumed to be the possible evaluations of a set of Boolean variables; the effects of the actions and the rewards of the states are represented in some succinct form. The succinct representation considered in this paper uses circuits (Mundhenk, Goldsmith, Lusena, & Allender, 2000), but others exist: decision trees, stochastic **STRIPS** operators, two-stage Bayes networks, variants of BDDs (Boutilier, Dean, & Hanks, 1999; Littman, 1997; Dean & Kanazawa, 1989; Dearden & Boutilier, 1997; Hansen & Feng, 2000). Littman (1997) has shown that these representations can be polynomially reduced to each other: the choice of circuits is motivated by their ease of use.

Typically, scenarios that can be expressed by MDPs can be informally represented in an amount of space that is on the same scale as the number of variables. For example, the domain in which there are ten coins and the ten actions of tossing them can be represented very succinctly: it is only necessary to specify that action $a_i$ tosses coin $i$ (e.g., the result of $a_i$ is a state in which the side the coin is on is head or tail with probability 0.5 and all other state variables are unchanged.) On the other hand, the explicit representation of this domain contains the set of all $2^{10}$ states and the representation of the transition function, which in turns requires the specification of the probability of each action $a$ to change a state $s$ into a state $s'$ for each pair of the $2^{10}$ states and each of the ten actions. As a result, the transition function contains a set of $2^{10}2^{10}10 = 10 \cdot 2^{20}$ probabilities.

Succinct representations, on the other hand, follow the intuition that a formal representation should not be too much larger than an informal representation of the same domain. While explicit representations are always exponential in the number of state variables, an informal representation may be very short: if this is the case, it is often also the case that such an informal description can be converted into a formal succinct one that is not too large.

Explicit and succinct representations lead to very different computational properties: for example, some problems (e.g., checking the existence of a policy of a given expected reward) are PSPACE-hard in the succinct representations (Mundhenk et al., 2000) but polynomial in the explicit one (Papadimitriou & Tsitsiklis, 1987). This apparent simplification is only due to the fact that complexity is measured relative to the size of the input of the problem, and the explicit representation introduces an artificial blow-up of this size.

In this paper, MDPs are assumed to be in a succinct representation. In particular, states are possible evaluations of a set of Boolean variables (state variables); the effects of the actions and the reward function are described using circuits. Since the number of states is exponential in the number of state variables, and a policy indicates the action to execute in each state, an explicit representation of policies is always exponential in the number of state variables. On the other hand, a succinct representation of an MDP may take only polynomial space in the number of state variables. If this is the case, the policy is exponentially larger than the MDP. However, as MDPs can be succinctly represented in a small amount of space, policies can be expressed in some succinct form as well (Koller & Parr, 1999). In this paper, policies are represented by circuits that take as input a state and (possibly) the history (the past states), and output the next action to execute.

The first result proved in this paper (Section 4) is that optimal policies, even in a succinct representation, may require an amount of space that is exponential in the size of the MDP.

This result is new for MDPs in succinct form; in particular, the hardness of finding an optimal policy does not imply anything about the policy size. Indeed, even in those cases in which finding an optimal policy is undecidable (Madani, Hanks, & Condon, 1999), the policy itself may be very short. Many hard problems, even some undecidable ones, are known to have very short solutions: for example, the solution of the halting problem is a single bit, but finding it is undecidable. Therefore, the impossibility of representing the solutions of a problem in polynomial space does not follow from the complexity of the problem.

Given that optimal policies cannot always be represented in space polynomial in the size of the MDP, a reasonable request is the best possible policy that can be represented within a given space bound (Section 5). We show that bounding the size of policies simplifies the policy existence problem. Bounding the size of the succinct representation of the value function (the function giving the expected reward of the states) further simplifies the problem (Section 6). This second bound is intended to shed light on the complexity of algorithms that work by estimating the expected reward of each state, such as the value iteration algorithm (Littman et al., 1995; Cassandra et al., 1997; Zhang & Zhang, 2001). We complete the analysis (Section 7) by considering the problem of finding policies of a given size and reward, when the size can be exponentially larger than the MDP. Implications and discussions of the results of this paper are given in the last section (Section 8).

## 2. Markov Decision Processes

Markov Decision Processes formalize problems of planning in probabilistic domains. Their components are: a set of states, a set of probabilistic actions, and a function that evaluates states according to a notion of goodness (reward function).

Formally, an MDP is a 5-tuple $\mathcal{M} = \langle \mathcal{S}, s_0, \mathcal{A}, t, r \rangle$, where: $\mathcal{S}$ is a set of states, $s_0$ is a distinguished state (the initial state), $\mathcal{A}$ is a set of actions, $t$ is a function representing the effects of the actions, and $r$ is a function giving the reward of the states.

The effects of the actions are not known for sure, but only according to a probability distribution. Therefore, the effects of actions cannot be represented using a function that maps a state into another state, given a specific action. The function $t$ is instead a function from actions and pairs of states to numbers in the interval $[0, 1]$. Such a function represents the probability of transitions: $t(s_1, s_2, a) = p$ means that the result of executing the action $a$ in the state $s_1$ is the state $s_2$ with probability $p$. The reward function is a measure of how much a state matches our goals. Formally, it is a function from states to integer numbers.

MDPs are assumed to be represented in a succinct form. Namely, the states are assumed to be represented by tuples of Boolean variables (state variables), i.e., the set of states is the set of propositional interpretations over this set of variables. The functions $t$ and $r$ are represented by Boolean circuits. While other representations are more commonly used in practice (e.g., probabilistic STRIPS operators, two-stage Bayes network, etc.) Boolean circuits have the advantage of being able to encode exactly all polynomial-time computable functions. This fact makes them very suitable for a computational analysis: indeed, if a transition or reward function is polynomial, then it can be encoded by a polynomial-size circuit without having to show the details of the encoding (Mundhenk et al., 2000; Boutilier et al., 1999). In order to encode some of the polynomial-time functions, on the other hand, the other representations require the introduction of new variables and dummy time points (Littman, 1997).

**Definition 1** *A succinct MDP is a 5-tuple $\mathcal{M} = \langle \mathcal{V}, s_0, \mathcal{A}, t, r \rangle$, where $\mathcal{V}$ is a set of Boolean variables, $s_0$ is a propositional interpretation over $\mathcal{V}$, $\mathcal{A}$ is a set of actions, $t$ is a circuit whose input is a pair of interpretations over $\mathcal{V}$ and an element of $\mathcal{A}$, and $r$ is a circuit whose input is an interpretation over $\mathcal{V}$.*

In a succinct MDP, the set of states is factored, i.e., it is the Cartesian product of the possible values of the variables. This is why succinct MDPs have also been called "factored MDPs". The term "succinct", however, is more appropriate, as $t$ and $r$ are not expressed as a product of something. It is also important to note that the set of actions $\mathcal{A}$ is explicitly represented: this affects the complexity of checking the consistency of a value function, i.e., Theorem 8.

A succinct MDP $\mathcal{M} = \langle \mathcal{V}, s_0, \mathcal{A}, t, r \rangle$ represents the (explicit) MDP $\mathcal{M}' = \langle \mathcal{S}, s_0, \mathcal{A}, t', r' \rangle$, where $\mathcal{S}$ is the set of propositional interpretations over variables $\mathcal{V}$; the transition function $t'$ is the function represented by the circuit $t$; the reward function $r'$ is the function represented by the circuit $r$. In other words, the value of $t'(s, s', a)$ is the output of the circuit $t$ when its inputs are $s$, $s'$, and a Boolean representation of $a$. The same applies to $r$ and $r'$.

Planning in deterministic domains consists of finding a sequence of actions to reach a goal state. Nondeterminism introduces two complications: first, the extent to which the goal is reached can only be probabilistically determined; second, the state at some time point cannot be uniquely determined from the initial state and the actions executed so far.

Planning in nondeterministic domains consists of finding the actions that lead to the best possible states (according to the reward function). Since the effects of actions are not known for sure, only an expected value of the reward can be determined. For example, if the result of applying $a$ in the state $s_0$ is $s_1$ with probability $1/3$ and $s_2$ with probability $2/3$, then the expected reward of executing $a$ is given by $r(s_0) + 1/3 \cdot r(s_1) + 2/3 \cdot r(s_2)$, since $r(s_0)$, $r(s_1)$, and $r(s_2)$ are the rewards of $s_0$, $s_1$, and $s_2$, respectively. Formally, the expected undiscounted reward is considered. This is the sum of the reward of each state weighted by the probability of reaching it.

The second effect of nondeterminism is that the best action to execute depends on the current state, which is not unambiguously determined from the initial state and the actions executed so far, as the effects of the actions are only probabilistically known. For example, executing $a$ may lead to state $s_1$ or $s_2$. After $a$ is executed, the actual result is known. At this point, it may be that the best action to execute in $s_1$ is $a'$, and is $a''$ in $s_2$: the optimal choice depends on the current state, which cannot be unambiguously determined from the initial state and the previous actions. In the simplest case, a policy is a function that gives the best action to execute in each state. Such policies are called stationary. A policy may also depend on the past states; such policies are called history dependent.

The reward associated with a policy is the expected average reward obtained by executing, in each state, the associated action. The horizon is assumed to be finite, i.e., only what happens up to a given number of steps $T$ is considered. The complexity of the problem changes according to whether the horizon $T$ is in unary or binary notation (Mundhenk et al., 2000). Informally, the assumption that the horizon is in unary notation means that the number of steps to consider is a polynomial in the size of the instance of the problem.

In this paper, $T$ is assumed in unary notation. This assumption has been called "short horizon" or "polynomial horizon" in other papers.

The size of the explicit representation of every policy for a succinct MDP is exponential in the number of state variables. However, some policies take less space in succinct representations. In this paper, the succinct representation employing circuits is considered: their input is the current state and (possibly) the history; their output is the next action to execute. An example of a policy with a small succinct representation is that of always executing the same action; such a policy is exponential in the explicit representation it is necessary to specify an action for each state and the number of states is exponential in the number of state variables.

The first question considered in this paper is whether it is always possible to represent the optimal policy of a succinct MDP with a circuit of size polynomial in the size of the MDP. Namely, a succinct policy is defined as a circuit that outputs, given the current state, an action.

**Definition 2** *A succinct stationary policy $P$ is a circuit that takes a state as input and outputs the action to execute in the state.*

A succinct policy is a circuit that represents a function from states to actions. Since (non-succinct) policies are functions from states to actions, succinct and non-succinct policies are in correspondence. The expected reward of a succinct policy and its optimality can therefore be defined in terms of the analogous concepts for non-succinct policies.

Succinct history-dependent policies are defined in a similar way: they are circuits from sequences of states to actions (such a representation is possible because of the finite horizon.) The expected reward and the optimality of a succinct history-dependent policy are defined from the corresponding concepts for non-succinct policies.

The first result of this paper is that, if all MDPs have optimal succinct policies of polynomial size, then the polynomial hierarchy coincides with its complexity class $\Pi_2^p$ (considered unlikely). The proof is based on compilability classes and reduction, which are summarized in the next section.

## 3. Complexity, Compilability, and Circuits

The reader is assumed to be familiar with the complexity classes P, NP, and the other classes of the polynomial hierarchy (Stockmeyer, 1976; Garey & Johnson, 1979). Some counting classes are also used in this paper: PP is the class of the problems that can be polynomially reduced to the problem of checking whether a formula is satisfied by at least half of the possible truth assignments over its set of variables. An alternative definition is that a problem is in PP if and only if it can be polynomially reduced to that of checking whether $\sum V(M) \geq k$, where $k$ is an integer, $V$ is a function from propositional interpretations to integers that can be calculated in polynomial time, and the sum ranges over all possible propositional interpretations (Johnson, 1990). The class NP$^{PP}$ is defined in terms of oracles: it is the class of problems that can be solved in polynomial time by a non-deterministic Turing machine that has access to a PP-oracle, which is a device that can solve a problem in PP in one time unit.

Without loss of generality, instances of problems are assumed to be strings. The length of a string $x \in \Sigma^*$ is denoted by $||x||$. The cardinality of a set $S$ is instead denoted by $|S|$.

A function $g$ is called *poly-time* if there exists a polynomial $p$ and an algorithm $A$ such that, for all $x$, the time taken by $A$ to compute $g(x)$ is less than or equal to $p(||x||)$. A function $f$ is called *poly-size* if there exists a polynomial $p$ such that, for all strings $x$, it holds $||f(x)|| \leq p(||x||)$; whenever the argument of such a function $f$ is a number, is is assumed to be in unary notation. These definitions extend to functions with more than one argument as usual.

Circuits are defined in the standard way. Whenever $C$ is a circuit and $s$ is a possible assignment of its input gates, its output is denoted by $C(s)$ its output. The formal definition of circuits is not important in this paper, as we take advantage of a well-known result in circuit complexity that relates poly-time functions with poly-size circuits. Informally, given a circuit and the value of its inputs, we can determine its output in polynomial time. A similar result holds in the other way around, that is, poly-time functions can be represented by means of circuits (Boppana & Sipser, 1990).

Formally, however, a poly-time function from strings to strings has a string of arbitrary length as argument, and a string of arbitrary length (in general) as output. Even considering only functions that have a binary output (i.e., a single bit), the input may be arbitrarily long. On the other hand, each circuit has a specified number of input gates. This is why the correspondence between poly-time functions and poly-size circuits is not one-to-one. However, the following correspondence holds: for any poly-time function from strings to strings, there exists a uniform family of poly-size circuits $\{C_0, C_1, C_2, \ldots\}$, where $C_i$ is a circuit with $i$ input gates that calculates the result of $f$ on all strings of length $i$; "uniform" means that that there exists a function that calculates $C_i$ from $i$ and runs in time that is bounded by a polynomial in the value of $i$.

As a result, the class P can be also defined as the set of problems that can be solved by a uniform family of circuits. By replacing the assumption of the family to be uniform with that of $C_i$ being of size polynomial in the value of $i$, the definition gives the class P/poly.

In this paper, the problem of whether all succinct MDPs have an optimal succinct policy of polynomial size is considered. Note that, for any given MDP, its succinct policies are circuits that take as input a state (or, a sequence of at most $T$ states). As a result, a policy is a single circuit, not a family.

The question is given a (negative) answer using two different proofs, based on different techniques: the first one is based on compilability classes (Cadoli, Donini, Liberatore, & Schaerf, 2002), while the second one employs standard complexity classes. The compilability classes have been introduced to characterize the complexity of intractable problems when some preprocessing of part of the data is allowed. The problems characterized in this way are those having instances that can be divided into two parts: one part is *fixed* (known in advance) and one part is *varying* (known when the solution is needed.) The problem of determining the action to execute in a specific state has this form: the MDP (with the horizon) is the part known in advance, as it describes the domain; on the contrary, the state is only determined once the previous actions have been executed. Compilability classes and reductions formalize the complexity of such problems when the first part can be preprocessed.

As is common in complexity theory, only decision problems are considered, i.e., problems whose solution is a single bit. Such problems are usually identified with languages (sets of strings): a language $L \subseteq \Sigma^*$ represents the decision problem whose output is 1 for all $x \in L$ and 0 for all $x \notin L$.

The problems whose instances are composed of two parts can be formalized as languages of pairs (of strings): a *language of pairs* $S$ is a subset of $\Sigma^* \times \Sigma^*$. In order to characterize these problems, the *non-uniform compilability classes* have been introduced (Cadoli et al., 2002). These classes are denoted by $\|\!\!\sim\!\!C$ and read "nu-comp-C", where C is an arbitrary uniform complexity class, usually based on time bounds, such as P, NP, etc.

**Definition 3 ($\|\!\!\sim\!\!C$ classes)** $\|\!\!\sim\!\!C$ *is composed of all languages of pairs $S \subseteq \Sigma^* \times \Sigma^*$ such that there exists a poly-size function $f$ from pairs to strings to strings and a language of pairs $S' \in C$ such that, for all $\langle x, y \rangle \in \Sigma^* \times \Sigma^*$, it holds:*

$$\langle x, y \rangle \in S \text{ iff } \langle f(x, ||y||), y \rangle \in S'$$

A problem is in $\|\!\!\sim\!\!C$ if it reduces to one in C after a suitable polynomial-size preprocessing (compiling) step. Any problem $S \subseteq \Sigma^* \times \Sigma^*$ that is in C is also in $\|\!\!\sim\!\!C$ (i.e., $f(x, n) = x$ and $S' = S$). Preprocessing is useful if a problem in C is in $\|\!\!\sim\!\!C'$ and $C' \subset C$: in this case, preprocessing decreases the complexity of the problem. Such a reduction of complexity is possible for some problems (Cadoli et al., 2002).

The tool used for proving that such a decrease of complexity is not possible is the concept of hardness with respect to compilability classes, which is in turn based on a definition of reductions. Since these general concepts are not really needed in this paper, only the condition based on monotonic polynomial reductions is presented. This condition is sufficient to prove that a problem in $\|\!\!\sim\!\!NP$ is not in $\|\!\!\sim\!\!P$ unless $NP \subseteq P/poly$, which is currently considered unlikely.

Let us assume that $\langle r, h \rangle$ is a polynomial reduction from 3sat to a problem of pairs $S$, that is, $r$ and $h$ are poly-time functions and $\Pi$ is satisfiable if and only if $\langle r(\Pi), h(\Pi) \rangle \in S$. This pair $\langle r, h \rangle$ is a monotonic polynomial reduction if, for any pair of sets of clauses $\Pi_1$ and $\Pi_2$ over the same literals, with $\Pi_1 \subseteq \Pi_2$, it holds:

$$\langle r(\Pi_1), h(\Pi_1) \rangle \in S \text{ iff } \langle r(\Pi_2), h(\Pi_1) \rangle \in S$$

Note that the second instance combines a part from $\Pi_2$ and a part from $\Pi_1$: this is intentional. Roughly speaking, such a reduction implies that the hardness of the problem $S$ comes from the second part of the instances only, as the first part $r(\Pi_1)$ of an instance that is the result of a reduction can be replaced by another one $r(\Pi_2)$ without changing its membership to $S$. If the complexity of a problem is due to a part of the instance only, preprocessing the other part does not reduce the complexity of the problem. The formal proof of the fact that the existence of such a reduction implies that $S$ is not in $\|\!\!\sim\!\!P$ (unless $NP \subseteq P/poly$) can be found elsewhere (Liberatore, 2001).

## 4. Super-polynomially Sized Policies

Suppose that it is always possible to find an optimal succinct policy $P$ of polynomial size. Since succinct policies are circuits by definition, deciding the action to execute in a state is a polynomial-time problem: given $s$ and $P$, just compute $P(s)$, the output of the circuit $P$ when $s$ is given as input. The whole two-step process of finding a policy and then using it to find the next action in a specific state can be seen as an algorithm for finding the next action to execute in the state. The first step (finding a policy) is likely to be very hard, but the second one is polynomial-time computable. This means that the problem of deciding the next action is compilable into $\Vdash$P. What is done in the present paper is to prove that this problem is instead not in $\Vdash$P. This implies that succinct policies cannot always be represented in polynomial space. To this aim, the formal definition of the problem of deciding the next action to execute is given.

**Definition 4** *The next-action problem is the problem of deciding, given an MDP, a horizon in unary notation, a state, and an action, whether the action is the one to execute in the state according to some optimal policy.*

The problem is proved NP-hard as an intermediate step. Actually, this result is easy to derive from known theorems: what is interesting is the reduction used in the proof. Namely, given a set of clauses $\Pi$, each clause being composed of three literals over a set of variables $X = \{x_1, \ldots, x_n\}$, an instance of the next-action problem in polynomial time can be built in polynomial time. This reduction is denoted by $f$; formally, this is a function from sets of clauses to quadruples $\langle \mathcal{M}, T, s, a \rangle$, where the first element is a succinct MDP, the second one is a number in unary (the horizon), the third one is a state, and the fourth one is an action. Let $L$ be the set of literals over $X$, and let $E = L \cup \{\mathsf{sat}, \mathsf{unsat}\}$. The MDP $\mathcal{M}$ is defined as:

$$\mathcal{M} = \langle \mathcal{V}, \epsilon, \mathcal{A}, t, r \rangle$$

The components of $\mathcal{M}$ are defined as follows.

**States:** The set of states is in correspondence with the set of sequences of at most $(2n)^3 + n + 1$ elements of $E$; this is obtained by using the following set of variables:

$$\mathcal{V} = \{q_i \mid 1 \leq i \leq \log((2n)^3 + n + 1)\} \cup \{v_i^j \mid 1 \leq i \leq \log(|E|),\ 1 \leq j \leq (2n)^3 + n + 1\}.$$

The idea is that the variables $q_i$ represent the length of the sequence in binary, while the variables $v_1^j, \ldots, v_{\log(|E|)}^j$ represent the $j$-th element of the sequence;

**Initial state:** The initial state is the interpretation representing the empty sequence $\epsilon$;

**Actions:** $\mathcal{A}$ contains three actions $A$, $S$, and $U$, and one action $a_i$ for each $x_i$;

**Transition function:** The action $A$ does not change the current state if either $\mathsf{sat}$ or $\mathsf{unsat}$ belong to the sequence that is represented by the current state; otherwise, its effect is to randomly select (with equal probability) a literal of $L$ and to add it to the sequence representing the current state; the effect of $S$ and $U$ is to add $\mathsf{sat}$ and $\mathsf{unsat}$ to the sequence, respectively (these are deterministic actions); the actions $a_i$ change the state only if either $\mathsf{sat}$ or $\mathsf{unsat}$, but not both, belong to the sequence; if this is the case, $a_i$ adds either $x_i$ or $\neg x_i$ to the sequence, with the same probability;

**Reward function:** This is the most involved part of the MDP. Given a sequence of $3m$ literals of $L$, the following 3cnf formula is considered:

$$C(l_1^1, l_2^1, l_3^1, \ldots, l_1^m, l_2^m, l_3^m) =$$
$$\{l_1^1 \vee l_2^1 \vee l_3^1, \ldots, l_1^m \vee l_2^m \vee l_3^m\}.$$

Given that the number of possible distinct clauses over $L$ is less than $(2n)^3$, any set of clauses can be represented as a sequence of $3m$ literals, where $m = (2n)^3$. The function $C$ encodes all sets of clauses over $L$ as sequences of literals.

The only sequences having reward different than zero are those composed of a sequence $s$ of $3m$ literals of $E$, followed by either sat or unsat, followed by a sequence $s' = l_1, \ldots, l_n$, where each $l_i$ is either $x_i$ or $\neg x_i$. Namely, the sequence $(s, \mathsf{unsat}, s')$ has reward 2; the sequence $(s, \mathsf{sat}, s')$ has reward 1 if the set of clauses $C(s)$ is not satisfied by the only model of $s'$, and $2^{n+1}$ otherwise;

Note that most of the states have reward 0. While the expected reward is calculated over all reached states, $r$ is defined in such a way that, if a state has nonzero reward, then all previous and succeeding states have reward zero. The reward function $r$ is defined this way for the sake of making the proof simpler; however, the expected reward is calculated over all states, including "intermediate" ones.

This MDP has a single optimal policy: execute $A$ for $3m$ times, then execute either $U$ or $S$, and then execute $a_1, \ldots, a_n$. The choice between $U$ and $S$ that gives the greatest expected reward depends on the result of the execution of $A$. Namely, each possible result of the execution of the first $3m$ actions corresponds to a set of clauses. The next action of the optimal policy is $U$ if the set is unsatisfiable and $S$ if it is satisfiable.

This is the definition of the MDP $\mathcal{M}$. The instance of the next-action problem is composed of an MDP, a horizon in unary, a state, and an action, and the problem is to check whether the action is optimal in the state. The horizon we consider is $T = (2n)^3 + n + 1$, the state is the one $s$ corresponding to the sequence of literals such that $C(s) = \Pi$, and the action is $S$. It is not possible to prove that the function $f$ defined by $f(\Pi) = \langle \mathcal{M}, T, s, S \rangle$ is a reduction from satisfiability to the next-action problem.

**Theorem 1** *If $f(\Pi) = \langle \mathcal{M}, T, s, S \rangle$, then $\mathcal{M}$ has an unique (stationary or history-dependent) optimal policy w.r.t. the horizon $T$, and $\Pi$ is satisfiable if and only if $S$ is the action to execute from $s$ in the optimal policy of $\mathcal{M}$ with horizon $T$.*

*Proof.* All MDPs defined as above have policies with positive reward: the policy of executing $A^{3m} U a_1, \ldots, a_n$ has expected reward equal to 2, since all its leaves have reward 2 and all internal nodes have reward 0. Such a sequence of actions can be executed by a stationary policy because the history can be inferred from the current state.

The sequences of actions that end up in a state with a positive reward are very similar to each other. Indeed, they all begin with $3m$ times the action $A$. Then, either $U$ or $S$ are executed, followed by the sequence $a_1, \ldots, a_n$. The only difference between these sequences is this choice between $U$ or $S$.

The optimal policies therefore execute $A$ in the first $3m$ time points, regardless of the generated state. They then execute either $U$ or $S$, and then execute $a_1, \ldots, a_n$. The choice between $U$ and $S$ can be made differently in different states: a policy can execute $U$ or $S$ depending on the state that results from the $3m$ executions of $A$.

sequence 1: $\underbrace{A \ldots A}_{3m} \ U \ a_1 \ \ldots \ a_n$

sequence 2: $\underbrace{A \ldots A}_{3m} \ S \ a_1 \ \ldots \ a_n$

Figure 1: Sequences leading to a state with reward $> 0$. All their fragments and extensions give reward 0.

Let us now consider the state after the $3m$ executions of $A$. Since each execution of $A$ generates a random literal, at this point the state is a sequence of $3m$ literals. Such a sequence represents the set of clauses that is later used by the reward function. Intuitively, at this point the optimal policy should execute $U$ if the set of clauses is unsatisfiable, and $S$ if it is satisfiable.

Let $s$ be a state that results from the execution of $A$ for $3m$ times, and let $C(s)$ be the corresponding formula. The expected reward of $s$ in the policy that executes the sequence $U, a_1, \ldots, a_n$ is 2 because this sequence leads to states that have all reward 2. On the other hand, the reward of $s$ when the policy executes $A, a_1, \ldots, a_n$ depends on the satisfiability of the formula represented by the state. Namely, this sequence leads to a state for each possible truth interpretation over $X$; the reward of such a state is 1 if $C(s)$ is not satisfied by the model, and $2^{n+1}$ otherwise. This means that the reward of $s$ in this policy is 1 if the formula is unsatisfiable, and at least $(2^n - 1)/2^n + 2^{n+1}/2^n = (2^n - 1)/2^n + 2$ if the formula has at least one model.

The optimal choice in $s$ is therefore $U$ if the formula $C(s)$ is unsatisfiable (reward 2) and $S$ otherwise (reward greater than or equal to $(2^n - 1)/2^n + 2^{n+1}/2^n > 2$). Since the history can be inferred from the state, the optimal choice does not depend on whether stationary or history-dependent policies are considered. □

Incidentally, this theorem implies that choosing the next action optimally is NP-hard. What is important, however, is that the function $f$ can be used to prove that the problem of choosing the next action optimally cannot be simplified to P thanks to a preprocessing step working on the MDP and the horizon only. This, in turn, implies the nonexistence of a polynomially sized circuit representing the optimal policy.

A polynomial reduction from 3sat to a problem does not necessarily prove that the problem cannot be efficiently preprocessed. Consider, however, the case in which the problem instances can be divided in two parts. The reduction itself can be decomposed in two separate functions, one generating the part of the instance that can be preprocessed and one generating the rest of the instance. In our case, $\mathcal{M}$ and $T$ form the part that can be preprocessed, while the state $s$ and the action $a$ are the rest of the instance. As a result, if $f(\Pi) = \langle \mathcal{M}, T, s, a \rangle$, the two functions are defined by:

$$r(\Pi) \ = \ \langle \mathcal{M}, T \rangle$$

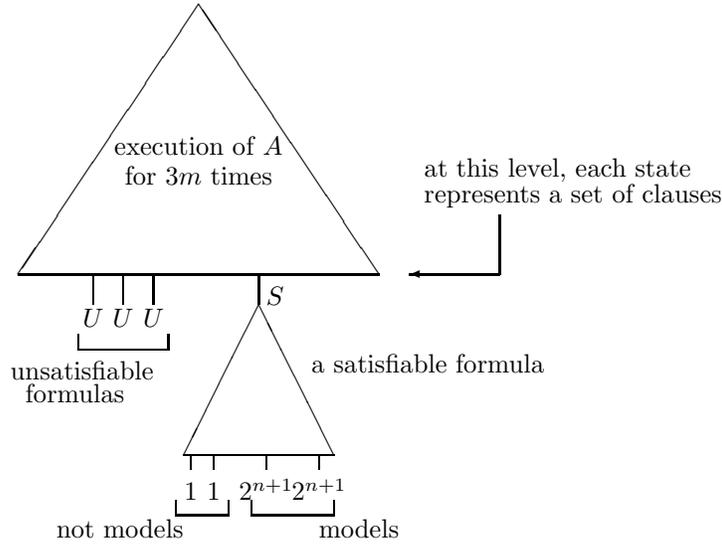

Figure 2: The optimal policy of the MDP of the proof.

$$h(\Pi) \;=\; \langle s, a \rangle.$$

Provided that the polynomial hierarchy does not collapse, a problem is not in $\|{\sim}\mathrm{P}$ if there exists a monotonic Polynomial reduction from 3sat to the problem (Liberatore, 2001). A polynomial reduction is monotonic if the two functions it is made of satisfy the following condition: for every pair of sets of clauses $\Pi_1$ and $\Pi_2$ of three literals over the same set of variables, if $\Pi_1 \subseteq \Pi_2$ then $\langle r(\Pi_1), h(\Pi_1)\rangle$ is a "yes" instance if and only if $\langle r(\Pi_2), h(\Pi_1)\rangle$ is a "yes" instance. Note that the second instance is $\langle r(\Pi_2), h(\Pi_1)\rangle$, that is, it combines a part derived from $\Pi_2$ and a part from $\Pi_1$.

The specialization of this condition to the case of the next-action problem is as follows. Let $\langle \mathcal{M}_1, T \rangle = r(\Pi_1)$ and $\langle \mathcal{M}_2, T \rangle = r(\Pi_2)$: the two horizons are the same because $\Pi_1$ and $\Pi_2$ are on the same alphabet. Let $\langle s, a \rangle = h(\Pi_1)$. Monotonicity holds if, for any two sets of clauses $\Pi_1$ and $\Pi_2$ over the same set of literals with $\Pi_1 \subseteq \Pi_2$, $a$ is the optimal action to execute in the state $s$ for $\mathcal{M}_1$ and $T$ if and only if it is so for $\mathcal{M}_2$ and $T$. The reduction $f$ defined above satisfies this condition.

**Theorem 2** *The function $f$ from sets of clauses to quadruples $\langle \mathcal{M}, T, s, a \rangle$ is a monotonic polynomial reduction.*

*Proof.* Let $\Pi_1$ and $\Pi_2$ be two sets of clauses over the same set of variables, each clause being composed of three literals, and let $\mathcal{M}_1$ and $\mathcal{M}_2$ be their corresponding MDPs. Since the MDP corresponding to a set of clauses depends—by construction—on the set of variables only, $\mathcal{M}_1$ and $\mathcal{M}_2$ are exactly the same MDP. Since the horizons of $r(\Pi_1)$ and $r(\Pi_2)$ are the same, we have $r(\Pi_1) = r(\Pi_2)$. As a result, for any state $s$ and action $a$, the latter is the optimal action to execute in $\mathcal{M}_1$ for the horizon $T$ if and only if it is so for $\mathcal{M}_2$, and this is the definition of monotonicity for the case of MDPs. □

This theorem implies that the next-action problem is hard for the compilability class $\Vdash\!\!\leadsto$NP. In turn, this result implies that some MDPs do not have any optimal succinct policy of size polynomial in that of the MDP and the horizon. More specifically, there is no way for storing the optimal actions for all states in such a way the required space is polynomial and the time needed to determine the action to execute in a state is polynomial as well.

**Theorem 3** *If there exists a data structure, of size polynomial in that of the MDP and the horizon, that allows computing the best action to execute (either from the current state or from the history) for any given MDP and horizon in unary notation in polynomial time, then* NP $\subseteq$ P/poly.

*Proof.* If such a data structure exists, the next-action problem is in $\Vdash\!\!\leadsto$P: given the fixed part of the problem only (the MDP and the horizon), it is possible determine such a data structure in the preprocessing step; the result of this step makes determining the next action a polynomial task. As a result, the next-action problem is in $\Vdash\!\!\leadsto$P. On the other hand, the existence of a monotonic polynomial reduction from propositional satisfiability to the next-action problem implies that if this problem is in $\Vdash\!\!\leadsto$P, then $\Vdash\!\!\leadsto$NP=$\Vdash\!\!\leadsto$P (Liberatore, 2001, Theorem 3). In turn, this result implies that NP $\subseteq$ P/poly (Cadoli et al., 2002, Theorem 2.12). $\square$

Note that NP $\subseteq$ P/poly implies that $\Sigma_2^p = \Pi_2^p = $ PH, i.e., the polynomial hierarchy collapses to its second level (Karp & Lipton, 1980). As a result, the above theorem implies that one can always represent the optimal policies in space polynomial in that of the MDP and the horizon only if the polynomial hierarchy collapses. Since the succinct representation of policies based on circuits is a subcase of data structures allowing the determination of the next action in polynomial time, the following corollary holds.

**Corollary 1** *If there exists a polynomial $p$ such that every succinct MDP $\mathcal{M}$ with horizon $T$ has succinct optimal policies (either stationary or history dependent) of size bounded by $p(||\mathcal{M}|| + T)$, then* NP $\subseteq$ P/poly *and* $\Sigma_2^p = \Pi_2^p = $ PH.

## 5. Finding and Evaluating Succinct Policies

The problem considered in this section is that of checking the existence of succinct policies of a given size and reward.

A subproblem of interest is that of evaluating a policy, that is, calculating its expected reward. While Mundhenk et al. (2000) found the complexity of this problem for various cases, they left open the case of full observability and succinct representation, which is the one considered in this paper. This problem has instead been analyzed by Littman, Goldsmith, and Mundhenk (1998) considering the succinct representation of plans based on the ST plan representation (which is also equivalent to other succinct representation.) The proof presented in the current paper could follow along similar lines.

Does the evaluation problem make sense, given that some MDPs do not have optimal succinct policies of polynomial size? From a theoretical point of view, super-polynomiality does not forbid a complexity analysis. Indeed, complexity is measured relative to the total size of the problem instances; the instances of the policy evaluation problem include both the policy and the MDP. If a policy is exponentially larger than the MDP, this means that the MDP is only a logarithmic part of the instance.

Formally, the problem of policy evaluation is: given an MDP, a policy, and a number $k$, decide whether the expected reward of the policy is greater than $k$. This is a counting problem, as it amounts to summing up the evaluations that result from computing a polynomial function over a set of propositional models. Not surprisingly, it is in PP.

**Theorem 4** *Given a succinct MDP $\mathcal{M} = \langle \mathcal{V}, s_0, \mathcal{A}, t, r \rangle$, a horizon $T$ in unary notation, a succinct policy $P$ (either stationary or history dependent), and a number $k$, deciding whether the policy $P$ has expected reward greater than $k$ is in* PP.

*Proof.* The expected reward of the policy is a weighted sum of rewards of states. Let us consider the sequence of states $s_0, s_1, \ldots, s_d$. The probability of this sequence of being the actual history can be computed as follows: for each pair of consecutive states $s_i, s_{i+1}$ there is a factor given by the probability $t(s_i, s_{i+1}, a)$, where $a$ is the unique action that is chosen by the policy in state $s_i$ (or, when the history is $s_0, \ldots, s_i$.) Multiplying all these factors, the result is the probability of the whole sequence $s_0, s_1, \ldots, s_d$ to be the actual history up to point $d$. Since $P(s_i)$ denotes the output of the circuit $P$ when $s_i$ is its input, then $P(s_i)$ represents the action that is executed in the state $s_i$. As a result, the probability of $s_0, s_1, \ldots, s_d$ of being the actual sequence of states is as follows:

$$H(s_0, s_1, \ldots, s_d) = \prod_{i=0,\ldots,d-1} t(s_i, s_{i+1}, P(s_i))$$

The same probability for history-dependent policies can be determined in the same way, but $P(s_i)$ is replaced by $P(s_0, \ldots, s_i)$. Given a specific sequence $s_0, \ldots, s_d$, it is possible to determine $H(s_0, \ldots, s_d)$ in time polynomial in the size of the sequence plus that of the MDP.

The expected reward of the policy can be calculated as the sum of the expected reward of each state $s_d$ multiplied by the probability of a sequence ending with $s_d$ being the actual history. This sum can be expressed as follows:

$$R(P) = r(s_0) + \sum_{s_1,\ldots,s_d \in \mathcal{S}}^{d=1,\ldots,T} H(s_0, \ldots, s_d) \cdot r(s_d).$$

The number $d$ ranges from 1 to $T$ to take into account all sequences up to length $T$. This is how "intermediate" states of the sequences are dealt with: for each sequence $s_0, \ldots, s_i, \ldots, s_d$, the sum above contains a term for each subsequence $s_0, \ldots, s_i$ as well.

Roughly speaking, the membership in PP is due to the fact that the expected reward of a policy is the result of a sum of an exponential number of terms, and each of those terms can be determined in polynomial time in a uniform manner. Formally, what is proved is that the problem can be expressed as the sum of terms $V(M)$, where $V$ is a polynomial function and $M$ ranges over all propositional interpretations over a given alphabet. To complete the proof, therefore, what is needed is to encode each possible sequence $s_0, \ldots, s_d$ as a propositional model, with the constraint that $H(s_0, \ldots, s_d) \cdot r(s_d)$ can be determined from this model in polynomial time.

The employed encoding is the following one: the alphabet is $X_1 \cup \cdots \cup X_T \cup Y$, where each set $X_i$ is a set of variables in one-to-one correspondence with the variables of the MDP, and $Y$ is a set of variables of size $\log(T)$. A model $M$ represents the sequence whose length is given by the values of $Y$ and whose $i$-th state is given by the values of $X_i$.

What is only left to show is that the expected probability of each sequence can be determined in polynomial time *given the model that represents the sequence*. This is true because, given $\mathcal{M}$, it is possible to rebuild the sequence from the model in time linear in the size of the model, and then evaluate $H$. Since sequences can be represented by propositional interpretations, and the function giving the weighted reward of each interpretation is polytime, the problem is in PP. □

This theorem shows that the problem of policy evaluation for succinct MDPs and horizon in unary notation is in PP. The problem can also be proved hard for the same class.

**Theorem 5** *Given a succinct MDP $\mathcal{M}$, a horizon $T$ in unary notation, a succinct policy $P$, and a number $k$, deciding whether the expected reward of $P$ is greater than $k$ is* PP-*hard.*

*Proof.* This theorem is proved by reduction from the problem of checking whether a formula is satisfied by at least half of the possible truth assignments over its variables. Let $Q$ be a formula over the alphabet $X = \{x_1, \ldots, x_n\}$. We define the MDP $\mathcal{M} = \langle \mathcal{V}, s_0, \mathcal{A}, t, r \rangle$ in such a way that its states are in correspondence with the sequences of literals over $X$. The actions are $\{a_1, \ldots, a_n\}$, where each $a_i$ modifies the state by adding $x_i$ or $\neg x_i$ to the sequence represented by the state, with the same probability. The reward function assigns 1 to all sequences that represent models of $Q$, and 0 to all other sequences. Namely, if a sequence contains a variable twice, or does not contain a variable, its corresponding state has reward 0.

Let $P$ be the policy of executing $a_i$ in each state composed of $i - 1$ literals. The states that result from the application of this policy at time $i$ are consistent sets of literals on the alphabet $x_1, \ldots, x_{i-1}$. If $i < n - 1$, such a state has reward 0. Therefore, only the states at time $T = n - 1$ are relevant to the calculation of the expected reward of $P$. These states are sets of literals that represent models over the alphabet $X$; their probability of being the state at time $T = n - 1$ is $1/2^n$. Given that the reward of a state is 1 if the corresponding model satisfies $Q$ and 0 otherwise, the expected reward of $P$ is $m/2^n$, where $m$ is the number of models of $Q$. □

The very same proof can be used to prove that the problem of *finding* the expected reward of a policy is #P-hard, but only decision problems are considered in this paper. The above two theorems allow concluding that the problem is PP-complete.

**Corollary 2** *Given a succinct MDP, a horizon in unary notation, a succinct policy, and a number $k$, deciding whether the expected reward of the policy is larger than $k$ is* PP-*complete.*

Let us now turn to the problem of checking the existence of a policy of a given size and expected reward. The same problem without the size constraint is PSPACE-hard (Mundhenk et al., 2000). The above corollary indicates that the size bound allows for a guess-and-check algorithm having a slightly lower complexity.

**Theorem 6** *Given a succinct MDP, a horizon in unary notation, a size bound $z$ in unary notation, and a reward bound $k$, checking the existence of a succinct policy of size bounded by $z$ and expected reward greater than or equal to $k$ is* $\mathrm{NP}^{\mathrm{PP}}$*-complete.*

*Proof.* Membership is easy to prove: guess a circuit of size $z$ representing a policy and check whether its expected reward is greater than $k$. Note that $z$ being in unary is essential.

Hardness is proved by reduction from e-majsat, a problem defined by Littman et al. (1998) as follows: given a formula $Q$ over variables $X \cup Y$, decide whether there exists a truth assignment over $X$ such that at least half of the interpretations extending it satisfy $Q$. This problem is $\mathrm{NP}^{\mathrm{PP}}$-complete (Littman et al., 1998).

Given an instance of e-majsat, its corresponding MDP is defined as follows: states represent sequences of literals over $X \cup Y$ (we assume, w.l.o.g., that $|X| = |Y| = n$.) There is an action $a_i$ for each variable in $Y$. Each $a_i$ adds the literal $y_i$ or $\neg y_i$ to the sequence representing the current state with the same probability. Each variable $x_i$ is associated with the actions $b_i$ and $c_i$, which add $x_i$ and $\neg x_i$ to the state, respectively. The reward of a state is 1 only if it represents a sequence of the form $[\neg]x_1, [\neg]x_2, \ldots, [\neg]y_n$ and all its literals satisfy $Q$, that is, the sequence is a complete interpretation that satisfies $Q$.

A policy with expected reward greater than 0 can only be made of a sequence whose $i$-th element ($1 \leq i \leq n$) is either $b_i$ or $c_i$, and its $i+n$-th element ($1 \leq i \leq n$) is $a_i$. The expected reward of this policy over a horizon $T = 2n$ is 1 only if at least half of the completions of the model defined by the actions $b_i$ and $c_j$ satisfy $Q$. Therefore, the MDP has a policy of reward 1 if and only if $Q$ is in e-majsat. □

This result can be used for an alternative proof of the claim that not all MDPs admit polynomially sized optimal succinct policies. This proof is interesting because it employs a different technique, and because it is conditioned differently than the previous one.

**Theorem 7** *If all MDPs have optimal succinct policies (either stationary or history dependent) of size polynomial in the sum of the size of the MDP and the length of the horizon then* $\mathrm{PSPACE} \subseteq \mathrm{NP}^{\mathrm{PP}}$*.*

*Proof.* Checking the existence of a policy of a given expected reward, regardless of its size, is PSPACE-hard (Mundhenk et al., 2000). On the other hand, if all policies can be represented in polynomial space, this problem coincides with that of checking the existence of a policy with a given bound on its size, and the latter problem is in $\mathrm{NP}^{\mathrm{PP}}$. □

Theorem 7 is proved using the fact that the problems of evaluating a policy of a given size and deciding the existence of a policy have different complexity characterizations. The same technique has been used by Papadimitriou and Tsitsiklis (1987) for the case of POMDPs in the explicit representation.

Precisely, the proof is composed of the following sequence of statements:

1. Evaluating the expected reward of a succinct policy is $C_1$-complete, where $C_1$ is a complexity class;

2. Deciding the existence of a policy (with no size bound) with a given expected reward is $C_2$-complete, where $C_2$ is a complexity class;

3. If any policy could be succinctly represented in space polynomial in the size of the MDP and the value of the horizon, the second problem could be solved by guessing a policy and then evaluating it; since the policy to guess has size polynomial in the size of the instance of $B$, then $C_2 \subseteq NP^{C_1}$.

What makes the proof worthy is the (probable) falsity of the conclusion $C_2 \subseteq NP^{C_1}$. In other words, the same proof can be applied to prove that a given data structure is not always polynomially large if:

1. there is a problem $A$ that is $C_1$-complete;

2. there is a problem $B$ that is $C_2$-complete;

3. problem $B$ can be expressed as: "there exists a data structure satisfying problem $A$".

This proof schema is simply the generalization of the one above: the data structure is the succinct policy; the first problem is that of evaluating a succinct policy; the second problem is that of deciding the existence of a policy giving a given expected reward. By definition, an instance satisfying $B$ implies *the existence of a data structure* satisfying $A$. If it is possible to replace "the existence of a data structure" with "the existence of a polynomial-size data structure" in this sentence, then $C_2 \subseteq NP^{C_1}$, as $B$ can be solved by guessing that data structure and then checking whether $A$ is satisfied or not. If the conclusion $C_2 \subseteq NP^{C_1}$ is false, then there are instances satisfying $B$ that are not related to any data structure satisfying $A$ that have size polynomial in the size of the instance of $B$.

## 6. Bounding the Value Function

Bounding the policy size is motivated by the fact that we are only interested in policies that can actually be stored: a policy can be used only if its size is less than or equal to the available storage space. In this section, a similar constraint is considered, motivated by how some algorithms for MDPs work. Namely, programs based on the popular value iteration algorithm (Littman et al., 1995; Cassandra et al., 1997; Koller & Parr, 1999; Zhang & Zhang, 2001) work by finding the value function, which is the function giving the expected reward that can be obtained from each state by executing the actions according to a given policy.

**Definition 5** *The value function $E$ of an MDP $\mathcal{M} = \langle \mathcal{S}, s_0, \mathcal{A}, t, r \rangle$ with horizon $T$ and policy $P$ is the function that gives the expected reward of a state $s$ at $i$ steps before the horizon:*

$$E(s, i) = \begin{cases} r(s) & \text{if } i = 0 \\ r(s) + \sum_{s' \in \mathcal{S}} t(s, s', P(s)) \cdot E(s', i - 1) & \text{otherwise.} \end{cases}$$

A similar definition can be given for history-dependent policies by including the history in the arguments of the value function and of the policy:

$$E(s_0, \ldots, s_j, i) = \begin{cases} r(s_j) & \text{if } i = 0; \\ r(s_j) + \sum_{s' \in \mathcal{S}} t(s_j, s', P(s_0, \ldots, s_j)) \cdot E(s', i - 1) & \text{otherwise.} \end{cases}$$

The value function for succinct MDPs is defined by simply replacing $s' \in \mathcal{S}$ with "$s'$ is a propositional interpretation over $\mathcal{V}$."

If the MDP and the policy are succinctly represented, the value function cannot necessarily be represented explicitly in polynomial space. Rather, some form of succinct representation is employed, usually by decomposition of the state space (for example, by grouping states with the same expected reward.) As it has already been done for policies, value functions are succinctly represented by circuits.

**Definition 6** *A succinct value function is a circuit $E$ whose inputs are a state $s$ and an integer $i$ and whose output is the expected reward of $s$ at $i$ points before the horizon.*

Given a policy $P$, there always exists an associated value function, which gives the expected reward that can be obtained from the state by executing the actions as specified by the policy. The converse, however, is not always possible. Some value functions, indeed, do not correspond to any policy. While any value function can be used to derive a policy by selecting the actions that maximize the expected reward of the resulting states (this is how value functions are often used), some value functions are completely unrelated to the actual expected reward of states. As an example, if the reward of $s_0$ is 0, and no action changes the states (i.e., $t(s, s, a) = 1$ for all actions $a$ and states $s$), then the value function $E$ such that $E(s_0, T) = 1000$ does not correspond to any policy. In other words, some value functions assign expected rewards to states in a way that is not consistent with the MDP, i.e., there is no way to obtain such a reward by executing whichever actions. Therefore, a value function may or may not be consistent, according to the following definition.

**Definition 7** *A value function $E$ is consistent with an MDP $\mathcal{M}$ and horizon $T$ if and only there exists a policy $P$ such that $E$ is the value function of $\mathcal{M}$, $T$, and $P$.*

An interesting property of value functions is that, in some cases, they actually represent policies. Indeed, a policy can be determined from a value function in polynomial time if the degree of non-determinism is bounded, and a list of possible states that may result from executing an action can be calculated in polynomial time. In particular, the set of states resulting from applying $a$ in $s$ are assumed to be the result of a circuit $n_a$.

**Definition 8** *A bounded-action MDP is a 6-tuple $\mathcal{M} = \langle \mathcal{V}, s_0, \mathcal{A}, \mathcal{N}, t, r \rangle$, where $\mathcal{M}' = \langle \mathcal{V}, s_0, \mathcal{A}, t, r \rangle$ is a succinct MDP and $\mathcal{N} = \{n_a\}$ is a set of circuits, one $n_a \in \mathcal{N}$ for each $a \in \mathcal{A}$, such that $n_a(s)$ is the list of states $s'$ such that $t(s, s', a) > 0$.*

Beside $\mathcal{N}$, the definition of bounded-action MDPs is the same as that of succinct MDPs. The difference can be explained in two ways: intuitively, it is assumed that the possible outcomes of actions can be determined in time polynomial in the size of the MDP; technically, the time needed to determine the possible states that result from applying an action is included in the size of the input.

The proof of $\text{NP}^{\text{PP}}$-hardness of the problem of policy existence only uses bounded-action MDPs, and therefore still holds in this case. This seems to contradict the intuition that a large degree of non-determinism is one of the sources of complexity of problems on MDPs. The next results will explain this contradiction.

Given a bounded-action MDP $\mathcal{M}$, a horizon, and a succinct value function $E$, we can identify in polynomial time a succinct policy that corresponds to $E$, that is, a policy that leads to the expected reward of states as specified by $E$. For the case of stationary policies, $P$ is determined as follows: for a given state $s$, we consider an action $a$ and a time point $i$, and check whether the result of executing $a$ is consistent with the value function, assuming that the time point is $i$. This is done by determining the sum of $t(s, s', a) \cdot E(s', i - 1)$. If this is equal to $E(s, i) - r(s)$, then the action $a$ is the action to execute, i.e., $P(s) = a$. The whole process is polynomial, that is, $P(s)$ can be determined in polynomial time.

If $E$ is consistent with an MDP, then there are policies for which the expected reward of each state $s$ is $E(s)$. As a result, a consistent value function $E$ represents a group of policies, all having the same expected reward. It therefore makes sense to consider the problem of finding $E$ rather than finding $P$. More precisely, since only decision problems are considered, the analyzed problem is that of checking whether there exists $E$ such that the expected reward of the corresponding policies is greater than or equal to a given number. Given $E$, the expected reward is simply given by $E(s_0, T)$. In order to check whether such a reward can actually be obtained from the MDP, however, we also have to check whether the value function $E$ is consistent with the MDP. This is the point where the assumption that the set of actions of a succinct MDP is not in a succinct representation is used.

**Theorem 8** *Checking whether a succinct value function $E$ is consistent with a bounded-action MDP and a horizon in unary is* coNP-*complete, both for stationary and history-dependent policies.*

*Proof.* The problem is to check whether there exists a policy $P$ that gives an expected reward as specified by $E$. Namely, for all $s$ and $i$, we have to check whether the equation in Definition 5 holds for some policy $P$. In turn, the existence of a policy means that, for each $s$, there exists an associated action $a$ that satisfies the equation when $P(s)$ is replaced by $a$.

Formally, let $\mathcal{M} = \langle \mathcal{V}, s_0, \mathcal{A}, t, r \rangle$ be the MDP, $T$ be the horizon, and $\mathcal{S}$ be the set of interpretations over the alphabet $\mathcal{V}$. The condition can be formally expressed as follows. For every state $s$, it holds that $E(s, 0) = r(s)$, and:

$$\forall s \in \mathcal{S} \;\; \exists a \in \mathcal{A} \;\; \forall i \in \{1, \ldots, T\} \; . \; E(s, i) = r(s) + \sum_{s' \in \mathcal{S}} t(s, s', a) \cdot E(s, i - 1).$$

This condition contains three alternating quantifiers; however, the second one ranges over $a \in \mathcal{A}$, while the third one ranges over $i = \{1, \ldots, T\}$. In both cases, the number of possibilities is polynomial in the size of the MDP and the horizon.

$$\forall s \; . \; \bigvee_{a \in \mathcal{A}} \; \bigwedge_{i = 1, \ldots, T} \left( E(s, i) = r(s) + \sum_{s' \in \mathcal{S}} t(s, s', a) \cdot E(s', i - 1) \right).$$

The number of terms $s' \in \mathcal{S}$ in the sum is not, in general, polynomial. The bounded action assumption, however, implies that the only states that are relevant are those $s'$ that belongs to $n_a(s)$, i.e., $s'$ is one of the elements of the list produced by the circuit $n_a$ when $s$ is given as its input.

$$\forall s \, . \, \bigvee_{a \in \mathcal{A}} \; \bigwedge_{i=1,\dots,T} \left( E(s,i) = r(s) + \sum_{s' \in n_a(s)} t(s, s', P(s)) \cdot E(s', i-1) \right).$$

Considering a given $s$ only, the condition can be checked in polynomial time. Since this condition have to be checked for all possible $s \in \mathcal{S}$, the problem is in coNP. The case of history-dependent policies is dealt with by replacing $\forall s$ with $\forall s_0, \dots, s_j$ and $i$ with $T - j$.

Hardness can be proved as follows: given a formula $Q$ over variables $X = \{x_1, \dots, x_n\}$, we build the succinct MDP with $\mathcal{V} = X$ and with a single action $a$. This action takes a number $i \in \{1, \dots, n\}$ with equal probability and changes the value of $x_i$. The reward function is 1 in a state if it satisfies $Q$, and 0 otherwise. The value function with $E(s, i) = 0$ for all $s \in \mathcal{S}$ and $i = 1, \dots, T$ is consistent with the MDP if and only if the formula $Q$ is unsatisfiable. Indeed, since the MDP only contains one action, the only possible stationary or history-dependent policy is that of always executing the action $a$. Such a policy leads to a random interpretation. The expected reward of this policy is 1 if and only if no interpretation satisfies $Q$. $\qquad \square$

Checking the existence of a consistent succinct value function of size bounded by $z$ and expected reward bounded by $k$ is therefore in $\Sigma_2^p$, as it can be done by guessing a succinct value function $E$, checking its consistency with the MDP and the horizon, and determining the expected reward of the initial state $E(s_0, T)$. Since consistency is in coNP for bounded-action MDPs, the problem is in $\Sigma_2^p$. The following theorem also shows that the problem is complete for this class.

**Theorem 9** *Given a bounded-action MDP $\mathcal{M}$, a horizon $T$ in unary notation, a size bound $z$ in unary notation, and a reward bound $k$, checking the existence of a succinct value function $E$ that is consistent with $\mathcal{M}$ and $T$, of size bounded by $z$ and expected reward bounded by $k$ is $\Sigma_2^p$-complete both for stationary and history-dependent policies.*

*Proof.* Membership: guess a succinct value function $E$ (i.e., guess a circuit with a state or history and an integer as input) of size at most $z$; check whether it is consistent with the MDP and the horizon, and whether $E(s_0, T) \geq k$. Since checking consistency in in coNP for both stationary and history-dependent policies, the problem is in $\Sigma_2^p$ in both cases.

Hardness is proved as in Theorem 6. We use the problem of checking whether there exists a truth evaluation over $X$ whose extensions to $X \cup Y$ are all models of $Q$, where $Q$ is a formula over $X \cup Y$ and $|X| = |Y| = n$.

The MDP that corresponds to $Q$ has sequences of literals over $X \cup Y$ as states. There is an action $a_i$ for each variable in $Y$ and two actions $b_i$ and $c_i$ for each variable in $X$. The effect of $a_i$ is to add either $y_i$ or $\neg y_i$ to the state, with the same probability. The actions $b_i$ and $c_i$ add $x_i$ and $\neg x_i$, respectively, and are therefore deterministic actions. The reward of a state is 1 if and only if the state is a sequence comprised of either $x_1$ or $\neg x_1$ followed by $x_2$ or $\neg x_2$, etc., and its literals form a model of $Q$. All other states have reward 0.

A policy with nonzero reward executes either $b_i$ or $c_i$ at time $i$, and then execute the actions $a_1, \dots, a_n$ in sequence. After the first $n$ actions, the state is exactly a model over $X$. The reward of the policy is the number of models that extend it and satisfy $Q$. Therefore, there exists a policy with reward 1 if and only if there exists a model over $X$ whose extensions all satisfy $Q$. ∎

This theorem shows that checking the existence of a succinct value function of a given size and reward is in a complexity class in the second level of the polynomial hierarchy for bounded-action MDPs. The corresponding problem for policies (instead of value functions) remains $\mathrm{NP^{PP}}$-hard for bounded-action MDPs. Assuming that $\Sigma_2^p \neq \mathrm{NP^{PP}}$, checking the existence of bounded-size succinct policies is harder than checking the existence of value functions, in the sense that more problems can be polynomially reduced to the latter problem than to the former. Let us consider this result.

1. The problem is easier because of the additional constraint on the size of the value function. Assuming $\Sigma_2^p \neq \mathrm{NP^{PP}}$, this implies that there are succinct policies of size polynomial in that of the MDP whose value function cannot be expressed as a circuit of size polynomial in that of the MDP, for bounded-actions MDPs. Translating succinct value functions into succinct policies is instead always feasible in polynomial space for bounded-actions MDPs. Therefore, any succinct policy can be translated into a succinct value function of polynomial size (in the size of the policy, the MDP, and the horizon), while the inverse translation is not always polynomial in size.

2. Given that the amount of physical memory is always bounded, the problem that is solved by algorithms finding the value function (such as value iteration) is actually that of checking the existence of a succinct value function of bounded size. This problem is easier than the similar problem with no bound, in the sense that the first problem can be polynomially reduced to the former but not vice versa, provided that $\Sigma_2^p \neq \mathrm{NP^{PP}}$.

Roughly speaking, finding a bounded-size succinct value function is simpler than finding a bounded-size succinct policy, but solving the former problem may produce solutions that are worse, in term of the overall reward, than the solutions of the latter.

## 7. Exponential Bounds

In the previous sections, the size bounds have been assumed to be in unary notation. This assumption has been made because, if it is not feasible to store the unary representation of the size, it is not feasible to store a data structure of that size as well. The unary notation formalizes the informal statement that the policy should take an amount of space that is polynomial in the size of the instance.

Let us now consider the problem of checking the existence of a policy of a bounded size and reward, where the bound on the reward is in binary notation: we are searching for a policy that can be exponentially larger than the MDP, but still of size bounded by a given number. This problem is of interest whenever an exponential succinct policy is acceptable, but still there is a limit on its size.

An important observation about circuits is that it is not necessary to consider circuits of arbitrary size. Indeed, any circuit with $n$ inputs and $m$ outputs is equivalent to a circuit of size $m2^n$. This is because any such circuit is equivalent to $m$ circuits, each having the same $n$ inputs and one output. Any such circuit represents a Boolean function, which can therefore expressed by a DNF in which each term contains all variables. Each such term represents a model: therefore, there are only $2^n$ different such terms if the alphabet is made of $n$ variables; as a result, the function can be expressed as a circuit of size $2^n$. As a result, any policy can be expressed as a circuit of size $|\mathcal{A}|2^n$, where $n$ is the number of state variables.

This fact has two consequences: first, it is not necessary to consider circuits of arbitrary size, as any policy can be represented by a circuit of size $|\mathcal{A}|2^n$; second, the problem of finding a succinct policy of size bounded by $z = |\mathcal{A}|2^n$ and bounded reward is the same as finding an arbitrary policy with the same bound on the reward. As a result, the problem of checking whether an MDP has a succinct policy of size bounded by $z$ and reward bounded by $k$ is at least as hard as the same problems with no bound on the size of the policy. The latter problem is PSPACE-hard; therefore, the former is PSPACE-hard as well. The same result holds for history-dependent policies: instead of $n$ input gates, there are $nT$ input gates. By this observation, the following result follows as a simple corollary of a result by Mundhenk et al. (2000).

**Corollary 3** *Checking whether an MDP and a horizon in unary notation have a succinct policy of expected reward greater than or equal to $k$, and having size bounded by $z$, is* PSPACE-*hard, if $z$ is in binary notation.*

Membership of this problem in PSPACE appears to be not so simple to prove. This is not surprising for the case of stationary policies, as the same question is still open for the problem with no bound on size. However, the same problem for history-dependent policies with no bound on size is known to be in PSPACE (incidentally, the problem with stationary policies but infinite horizon is EXPTIME-complete, as proved by Littman, 1997.) The proof of this result cannot however be modified to cover the addition of a bound on size. Intuitively, we are not only looking for a policy with a given reward, but also of a given size. The constraint on the reward is somehow easier to check, as it is done locally: the expected reward of a state is obtained by summing up the reward of the possible next states. On the other hand, the size of the circuit representing the policy cannot be checked until the whole circuit has been determined, and this circuit can be exponentially large.

What it is proved in this section is that the EXPTIME-hardness of the problem for history-dependent policies is related to an open conjecture in computational complexity (P=PSPACE). Namely, it is proved that P = PSPACE implies that the problem is in P. This result can be rephrased as: if the problem is not in P, then P $\neq$ PSPACE. Since EXPTIME-hard problems are not in P, if the problem is EXPTIME-hard, then P $\neq$ PSPACE. This conclusion is not really unlikely: on the contrary, it is believed to be true. On the other hand, proving the problem to be EXPTIME-hard is at least as hard as solving a conjecture that has been open for more than twenty years. As an intermediate step, it is proved that P = PSPACE implies the existence of polynomially-sized history-dependent policies for all MDPs.

**Theorem 10** *If* P = PSPACE, *then every succinct MDP with a horizon in unary notation has an optimal succinct history-dependent policy of size polynomial in the size of the MDP and the horizon.*

*Proof.* Let us consider a state that results from a sequence of actions. Since the considered policies are history dependent, it is possible to find the optimal choices from this point on by taking the current state as the new initial state of the MDP, reducing the horizon, and determining the optimal policy of this modified MDP.

Since the problem of checking whether an MDP has a policy of a given reward is in PSPACE (Mundhenk et al., 2000), it is in P by assumption. By binary search, it is possible to determine the expected optimal reward of an MDP in polynomial time. Since the expected optimal reward of each state can be computed by finding the expected optimal reward of an MDP, this problem is polynomial as well. The function that determines the optimal expected reward from a state can be therefore represented by a polynomial circuit.

The best action in a state is the one that leads to the best possible next states. It can be determined in polynomial space by checking, for each action, its possible next states, and determining their expected reward. Therefore, the best action to execute is in PSPACE, and is therefore in P by assumption. As a result, this optimal policy can be represented by a polynomial circuit. □

The following corollary is an immediate consequence.

**Corollary 4** *If* $P = PSPACE$, *then the problem of existence of a succinct history-dependent policy of a given expected reward and with a size bound in binary notation is in* $P$.

*Proof.* If $P = PSPACE$, any MDP has an optimal history-dependent policy of polynomial size. The problem can therefore be solved by iterating over all possible circuits whose size is bounded by this polynomial. The problem can therefore be solved with only a polynomial amount of memory, and is therefore in PSPACE. By assumption, it is in P as well. □

At a first look, this corollary seems to prove that the problem is in PSPACE. However, it does not prove such a result. This is shown by the following related example: a problem that is is $\Sigma_2^p$-complete is in P if $P = NP$; however, it is generally believed that $\Sigma_2^p$-complete problems are not in NP.

An easy consequence of this result is that the EXPTIME-hardness of the problem would imply that $P \neq PSPACE$. Indeed, if the problem is EXPTIME-hard then it is not in P, as these two classes are known to be different thanks to the theorem by Hartmanis and Stearns (1965). As a result, it cannot be that P=PSPACE, which has been proved to imply that the problem is in P. In other words, if the problem of history-dependent policy existence with a size bound in binary notation is EXPTIME-hard then $P \neq PSPACE$.

## 8. Conclusions

It is not always possible to represent the optimal policies of a succinct MDP using circuits of size polynomial in that of the MDP and the horizon. This result affects the choice of how policies are generated and executed. Indeed, planning in a nondeterministic scenario can be done in two ways:

1. Determine the actions to execute in all possible states all at once (i.e., determine the whole policy); in each state, the corresponding action is executed;

2. Determine the best action to execute in the initial state only; execute it and observe the resulting state; find the best action in the new state, etc.

Many algorithms for MDPs find a representation of the whole policy, and only a few solve the next-action problem directly (Kearns, Mansour, & Ng, 2002). Our result formally proves that the optimal policy cannot be always represented in polynomial space, unless the polynomial hierarchy collapses. This result holds not only for the existing algorithms (such as value iteration or policy iteration), but also for any other algorithm that finds the whole policy all at once.

While the second solution can be theoretically optimal, it involves finding the best action at each time step, and this problem is hard, as proved by Theorem 1. The advantage of the first solution is that the only hard step is to find the optimal policy; finding the action to execute in a state is then polynomial.

The impossibility of always representing the optimal policies in polynomial space raises a new problem: since the size of physical memory is bounded, it is not feasible to search for the best among all policies, but only among those it is possible to store, that is, those bounded in size by the available memory size. The problem of checking whether a succinct MDP has a succinct policy of a given size and reward is proved to be $NP^{PP}$-complete, and is therefore slightly easier than the same problem without a bound, which is PSPACE-hard (Mundhenk et al., 2000). A similar result has been proved for a slightly different formalization of non-deterministic planning by Littman et al. (1998).

This complexity result holds only if policies are represented in a particular succinct form, that of circuits giving the next action from the current state and (possibly) the history. Nevertheless, different representations of policies lead to different results. Namely, some algorithms actually find a value function (a function determining the expected reward in each state), which can be considered as a representation of a policy when the states that result from the execution of an action can be listed in polynomial time. In particular, algorithms based on value iteration, when applied with some form of decomposition, find a succinct representation of the value function. Finding a succinct representation of such a function, with a given bound on its size, has been proved easier than finding a succinct policy, when the states that result from the execution of an action can be listed in polynomial time: it is $\Sigma_2^p$-complete.

This result has two consequences: first, the problem these algorithms solve is in a smaller complexity class than the problem with no bound on size; second, some policies can be represented in polynomial space, while their associated value functions cannot: there exists a trade-off between complexity and ability of finding good solutions. This result is also interesting because it characterizes the complexity of a family of algorithms (those determining the value function in some succinct form) rather than the complexity of a problem. This result is therefore of a kind that is between the efficiency analysis of a single algorithm (e.g., the big-O running time) and the inherent complexity of the problem (e.g., the NP-completeness). As such, it is similar to results about complexity of proof procedures (Beame & Pitassi, 1998; Egly & Tompits, 2001). Our analysis is however limited to the case of exact value functions. Approximation is an interesting open problem. The effects of the various restrictions that can be done on the MDP are interesting open problems as well.

The analysis with a bound on size has been initially done assuming that the bound is polynomial in the size of the instance. This is justified by the fact that the resulting policy should not be too much larger than the MDP. However, a moderate increase may be tolerated. Therefore, we considered the same problem removing the assumption of polynomiality of the bound. This is a new problem, but is PSPACE-hard just as the problem with no bound on size is. However, it is not EXPTIME-hard unless $P \neq PSPACE$. This result shows that proving such hardness result is at least as hard as proving a conjecture that has been open for more than twenty years.

Let us now discuss how the results presented in this paper relate to similar ones in the literature. As already remarked in the Introduction, the complexity of problems related to finding a policy does not necessarily imply that policies cannot be compactly represented. Namely, even a result of undecidability does not forbid compactness of policies. Therefore, our result is not implied by previous complexity results in the literature. On the other hand, a result of non-polynomiality of the size of policies of POMDPs already appeared in the paper by Papadimitriou and Tsitsiklis (1987). Namely, they proved that, unless $PSPACE = \Sigma_2^p$, there is no algorithm $A$ and mapping $\mu$ from POMDPs in explicit form to strings of polynomial length that can be used by $A$ to compute the optimal action. This result basically proves the non-polynomiality of policies. However, it cannot imply ours, as it holds for POMDPs in the explicit representation with only non-positive rewards; the same problem that is PSPACE-hard in their formalization is polynomial in ours (Mundhenk et al., 2000). More precisely, the two results cannot be derived from each other. Some related results in the literature are about the "non-representability" (as opposite to the "compact-representability" studied in this paper): Littman (1997) has shown that (infinite horizon) plan existence is EXPTIME-complete, while Littman et al. (1998) have shown that the same problem restricted to looping plans is PSPACE-complete: as a result, infinite-horizon policies cannot be represented at all as looping plans unless EXPTIME=PSPACE.

During the review period of this paper, a different proof of the impossibility of representing the optimal policies of all MDPs in polynomial space has been published (Allender, Arora, Kearns, Moore, & Russell, 2002). This new proof improves over the ones presented in this paper and in its conference version (Liberatore, 2002) in two ways: first, the condition under which the new result holds is $PSPACE \nsubseteq P/poly$ instead of $NP \nsubseteq P/poly$ and $PSPACE \nsubseteq NP^{PP}$; second, the proof holds even for approximately optimal policies. The other differences (the new proof is for the infinite horizon with a discount factor, the reward function is linear, two-levels Bayes nets are used instead of circuits) are inessential, i.e., the proofs can be modified in such a way they are not affected by these other differences. The current paper also contains results on the problem with a bound on the policy or the value function.

Let us now consider the other complexity results about MDPs in the literature. The problem of deciding whether a succinct MDP with a horizon in unary notation has a policy of a given reward is PSPACE-hard, and is PSPACE-complete for history-dependent policies. The same problem with a bound in unary notation on the size of the policy is $NP^{PP}$-complete. This class contains the class $P^{PP}$, which in turn contains the whole polynomial hierarchy. Therefore, a problem that is $NP^{PP}$-complete is hard for any class of the polynomial hierarchy. This means that a bound on size in unary notation does not decrease the complexity of the problem much. On the other hand, bounding the size of the value function representation decreases the complexity more, as the problem is $\Sigma_2^p$-complete.

To conclude, observe that negative results (impossibility of polynomial policies and hardness results) hold for POMDPs, since MDPs are special cases of POMDPs in which everything is observable. Our results, however, only apply to the case in which both the POMDP and the policy are in succinct form. The case of explicit representation has been studied by Papadimitriou and Tsitsiklis (1987) (their results are discussed above), and by Mundhenk (1999), who considered the problem of deciding whether a POMDP in explicit form has a $c$-small policy, given $c$, where $c$-small-ness includes a bound on size depending on $c$.

## Acknowledgments

Many thanks to Michael Littman and the anonymous referees for their suggestions. Part of this work appeared in the proceedings of the Eighteenth National Conference on Artificial Intelligence (AAAI-2002).